\documentclass[10pt,a4paper,twocolumn]{article}

\usepackage{graphicx} 

\usepackage{dcolumn}           
\newcolumntype{.}   {D{.}{.}{-1}} 
\newcolumntype{d}[1]{D{.}{.}{#1}} 
\newcolumntype{e}   {D{E}{E}{-1}} 
\newcolumntype{E}[1]{D{E}{E}{#1}} 

\usepackage{amsmath}
\usepackage{mathptmx}

\usepackage{amssymb}
\usepackage{amsthm}

\usepackage{float}

\usepackage{setspace}

\usepackage{sectsty}

\newcommand{\myFontSize}{\fontsize{10}{12}\selectfont}
\renewcommand{\normalsize}{\fontsize{10}{12}\selectfont}
\sectionfont{\myFontSize}       
\subsectionfont{\rm\myFontSize\itshape} 

\usepackage{titlesec}
\titlespacing*{\section}{0pt}{10pt}{0pt}
\titlespacing*{\subsection}{0pt}{10pt}{0pt}

\usepackage{secdot}
\sectiondot{subsection}
\sectionpunct{section}{. }    
\sectionpunct{subsection}{. } 

\usepackage{caption}
\usepackage{subfig}
\usepackage{xcolor} 
\usepackage{fancyhdr} 
\usepackage{lastpage}
\renewcommand{\headrulewidth}{0pt} 

\usepackage[numbers]{natbib}

\usepackage[shortcuts,acronym,nonumberlist]{glossaries}
\makeglossaries 
\newacronym{CDM}{CDM}{conjunction data message}
\newacronym{TCA}{TCA}{time of closest approach}
\newacronym{LEO}{LEO}{low-Earth orbit}
\newacronym{SSN}{SSN}{Space Surveillance Network}
\newacronym{O/O}{O/O}{owners/operators}
\newacronym{ESA}{ESA}{European Space Agency}
\newacronym{RSO}{RSO}{resident space object}

\newacronym{LSTM}{LSTM}{long short-term memory}
\newacronym{DDPM}{DDPM}{denoising diffusion probabilistic model}
\newacronym{MAE}{MAE}{mean absolute error}
\newacronym{RMSE}{RMSE}{root mean squared error}

\begin{document}

%
\twocolumn[
\begin{@twocolumnfalse}

\vspace{0pt}
\begin{center}
    
    \vspace{15pt}
    \textbf{Predicting the Position Uncertainty at the Time of Closest Approach with Diffusion Models}
    
    \vspace{10pt}
    \textbf{
        \selectfont\fontsize{10}{0}\selectfont Marta~Guimarães~\textsuperscript{a,b*},~Cláudia~Soares~\textsuperscript{b},~Chiara Manfletti\textsuperscript{a} 
    }
\end{center}


\vspace{-10pt} 
\begin{flushleft}
    \textsuperscript{a}\textit{
    \fontfamily{ptm}\selectfont\fontsize{10}{12}\selectfont Neuraspace, Portugal}, \underline{\{marta.guimaraes, chiara.manfletti\}@neuraspace.com}
    \\
    \textsuperscript{b}\textit{
        \fontfamily{ptm}\selectfont\fontsize{10}{12}\selectfont FCT-UNL, Portugal},
        \underline{claudia.soares@fct.unl.pt}
    \\
    \textsuperscript{*}\fontfamily{ptm}\selectfont\fontsize{10}{12}\selectfont Corresponding Author  
\end{flushleft}

\begin{abstract}
The risk of collision between resident space objects has significantly increased in recent years. As a result, spacecraft collision avoidance procedures have become an essential part of satellite operations. To ensure safe and effective space activities, satellite owners and operators rely on constantly updated estimates of encounters. These estimates include the uncertainty associated with the position of each object at the expected \gls{TCA}. These estimates are crucial in planning risk mitigation measures, such as collision avoidance manoeuvres. As the \gls{TCA} approaches, the accuracy of these estimates improves, as both objects’ orbit determination and propagation procedures are made for increasingly shorter time intervals. However, this improvement comes at the cost of taking place close to the critical decision moment. This means that safe avoidance manoeuvres might not be possible or could incur significant costs. Therefore, knowing the evolution of this variable in advance can be crucial for operators. This work proposes a machine learning model based on diffusion models to forecast the position uncertainty of objects involved in a close encounter, particularly for the secondary object (usually debris), which tends to be more unpredictable. Diffusion models are a class of state-of-the-art deep learning probabilistic generative models based on non-equilibrium thermodynamics. They capture multiscale effects by creating a succession of simplified views of a sequence, modelled as a Markov chain. Such a Markov chain can be reversible, and in this mode the model develops complex and realistic predictions from noisy and partial information. Such properties are well-suited to predicting the position uncertainty of space objects at the \gls{TCA}. We compare the performance of our model with other state-of-the-art solutions and a naïve baseline approach, showing that the proposed solution has the potential to significantly improve the safety and effectiveness of spacecraft operations.

\noindent{{\bf Keywords:}} Diffusion Models, Forecasting, Deep Learning, Space Debris, Space Traffic Management \\

\end{abstract}

\end{@twocolumnfalse}
]


\section{Introduction}

With the increasing number of \glspl{RSO}~\cite{ESA2023} and the complexity of their interactions, innovative approaches for managing collision risks have become imperative~\cite{Kessler1978, Virgili2016}. Ensuring safe and effective space activities requires satellite \gls{O/O} to rely on constantly updated estimates of encounters, including the uncertainty associated with the position of each object at the expected time of \gls{TCA}. Accurate estimation of the \gls{TCA} is crucial for planning risk mitigation measures, such as collision avoidance manoeuvres. However, obtaining precise \gls{TCA} estimates often happens close to the critical decision moment, where safe avoidance manoeuvres may not be feasible or incur significant costs.

Given these challenges, the ability to forecast the evolution of position uncertainty in advance becomes paramount for satellite \gls{O/O}. Having insights into the evolution of this crucial variable enables operators to proactively prepare and strategise to minimise the risks associated with close encounters. In this context, this work proposes utilising a machine learning model based on diffusion models~\cite{Sohl-Dickstein2015} to forecast the position uncertainty of objects involved in close encounters, with a particular emphasis on secondary objects like debris, which tend to exhibit greater unpredictability. 

Diffusion models represent a cutting-edge class of deep learning probabilistic generative models grounded in non-equilibrium thermodynamics. These models excel at capturing multi-scale effects by generating a sequence of simplified perspectives, modelled as a Markov chain. The reversible nature of these Markov chains enables the model to generate complex and realistic predictions even when presented with noisy and partial information. Such characteristics make diffusion models well-suited for predicting the position uncertainty of space objects at the \gls{TCA}, allowing for improved decision-making and risk assessment.

\subsection{Related Work}
\begin{figure}[htb!]
    \centering
    \includegraphics[width=0.4\textwidth]{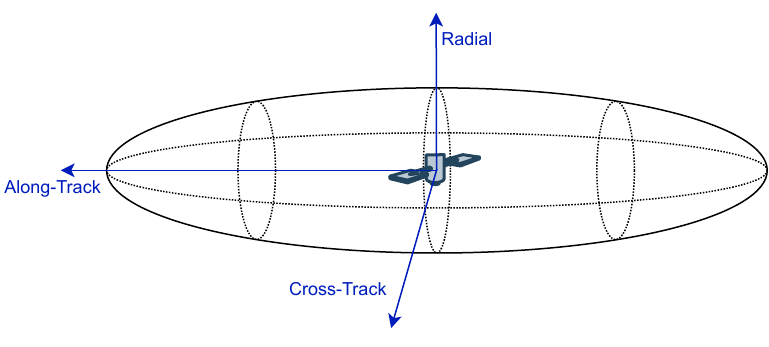}
    \caption{Satellite position uncertainty. In this example, the uncertainty ellipsoid is aligned with the satellite motion, i.e., with the satellite reference frame, represented in blue.}
    \label{fig:covariance}
\end{figure}

Due to the rapidly increasing number of resident space objects, there has been a growing interest in studying and forecasting the evolution of the object's position uncertainty~\cite{Virgili2019}. More concretely, the prediction target usually considered is the main diagonal element of the chaser covariance matrix, i.e., the position uncertainty/variance on the along-track axis (Figure \ref{fig:covariance}). Such a decision relies on the fact that the main eigenvector of the covariance matrix is typically almost aligned with the satellite motion. 

~\citet{Metz2020} was one of the first researchers that explored the use of machine learning to predict the position uncertainty at the \gls{TCA}. This work investigated the use of different strategies, namely, decision tree methods and neural networks, such as multilayer perceptron~\cite{Bishop2006} and \gls{LSTM}~\cite{Hochreiter1997} networks. Building upon this research,~\citet{Stroe2021} also leveraged \glspl{LSTM} but innovatively combined them with the attention mechanism~\cite{Vaswani2017}. Moreover, they investigated predicting at different time horizons, introducing new dimensions to the problem.

The problem of quantifying uncertainty associated with the predictions generated by machine learning models is crucial for their practical applications. Deep learning approaches, although powerful, often lack interpretability, making it challenging for operators to trust the outputs of neural networks. To address this concern,~\citet{Pinto2020} proposed a Bayesian deep learning approach by leveraging the Monte Carlo dropout technique applied to a stack of \glspl{LSTM}. Such an approach provides probability distributions for predictions, enabling robust uncertainty quantification. This allows satellite operators to make more informed decisions, considering the level of uncertainty associated with the model's predictions.

\subsection{Contributions}

In this study, we present a novel methodology that leverages the power of diffusion models for forecasting the position uncertainty of space objects in close encounters. By surpassing the limitations of conventional approaches and offering probabilistic uncertainty estimates, our research significantly advances the state-of-the-art in space situational awareness and collision risk assessment.

\section{Diffusion Models}

Generative models in machine learning have recently gained popularity, with diffusion models being a notable example. These models employ a forward diffusion process to introduce perturbations to the observed data, followed by a backward process that aims to recover the original data. In the forward process, noise is introduced at multiple stages, with the noise level varying at each step. The backward process involves several denoising steps that gradually eliminate the noise. A neural network is used to parameterise the backward process, enabling it to generate new samples from almost any initial data once it has been trained.

\subsection{Diffusion Models for Time Series Applications}

In recent years there has been a significant extension of diffusion models to various time series-related applications, including time series forecasting~\cite{Rasul2021, Li2022, Bilo2023}, time series imputation~\cite{Tashiro2021, Alcaraz2022, Liu2023}, and time series generation~\cite{Lim2023}.

In this work, we have built upon the work of RePaint~\cite{Lugmayr2022}, a \gls{DDPM} based inpainting approach, due to its efficiency and simplicity. Such a model was originally proposed in the scope of image generation and aimed to predict missing pixels of an image using a mask region as a condition.

\subsection{Denoising Diffusion Probabilistic Models}


\begin{figure*}[t!]
    \centering
    \includegraphics[width=\textwidth]{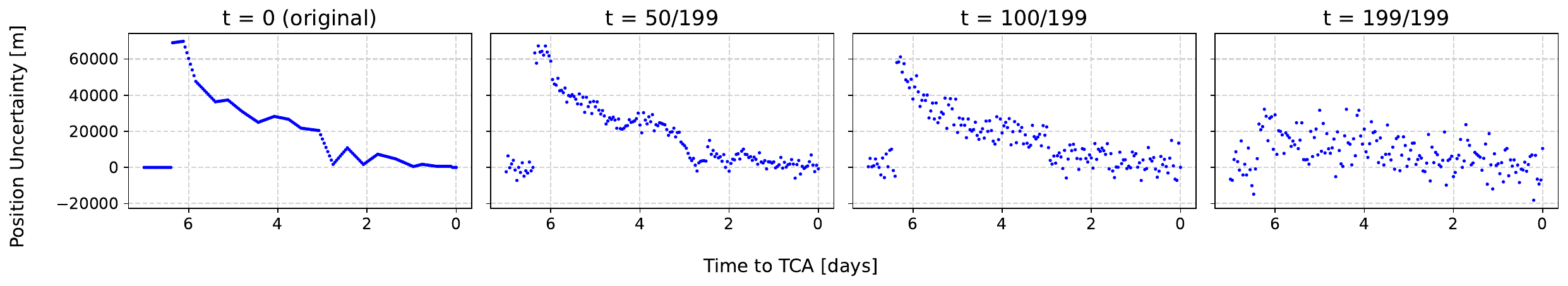}
    \caption{Forward diffusion process applied to a training sample with 200 diffusion steps.}
    \label{fig:forward_diffusion_process}
\end{figure*}


\Glspl{DDPM} implement the forward and backward processes through two Markov chains~\cite{Sohl-Dickstein2015, Ho2020}.

Given a data point from the observed data distribution $\textbf{x}_{0}\sim q(\textbf{x})$, the forward diffusion process adds a small amount of Gaussian noise to the sample in $T$ steps, producing a sequence of noisy samples $\textbf{x}_1, ..., \textbf{x}_T$. Each step is given by
\begin{equation}
    q\left(  \textbf{x}_t | \textbf{x}_{t-1} \right) = \mathcal{N}\left(\textbf{x}_t;\sqrt{1-\beta_t}\textbf{x}_{t-1}, \beta_t\textbf{I}  \right).
    \label{eq:forward_process}
\end{equation}
The sample $\textbf{x}_t$ is obtained by adding i.i.d.\ Gaussian noise with variance $\beta_t$ at time step $t$ and scaling the previous sample $\textbf{x}_{t-1}$ with $\sqrt{1-\beta_t}$ according to a variance schedule $\left\{ \beta_t\in (0,1) \right\}^{T}_{t=1}$.

The \gls{DDPM} is trained to reverse the process in~\eqref{eq:forward_process}, which is modelled by a neural network that predicts the parameters $\mu_\theta(\textbf{x}_t, t)$ and $\Sigma_\theta(\textbf{x}_t, t)$ of a Gaussian distribution,
\begin{equation}
   \label{eq:ddpm_training}
   p_\theta\left(  \textbf{x}_{t-1} | \textbf{x}_{t} \right) = \mathcal{N}\left(\textbf{x}_{t-1}; \mu_\theta(\textbf{x}_t, t), \Sigma_\theta(\textbf{x}_t, t)  \right)
\end{equation}

A convenient property of such a process is that we can sample $\textbf{x}_t$ at any time step $t$ in a closed form using a reparameterisation trick. Let $\alpha_t = 1 - \beta_t$ and $\bar{\alpha}_t=\prod_{i=1}^{t}\alpha_i$, we can thus rewrite~\eqref{eq:forward_process}, as a single step,
\begin{equation}
    \label{eq:sample_forward}
    q_\theta\left(  \textbf{x}_{t} | \textbf{x}_{0} \right) = \mathcal{N}\left(\textbf{x}_{t}; \sqrt{\bar{\alpha}_t}\textbf{x}_0, (1-\bar{\alpha}_t) \textbf{I} \right).
\end{equation}

By doing so, we can train the \gls{DDPM} by sampling $\textbf{x}_t$ and corresponding noise that is used to transform $\textbf{x}_0$ to $\textbf{x}_t$.

\subsection{Inpainting}
Assuming a trained unconditional \gls{DDPM},~\eqref{eq:ddpm_training},~\citet{Lugmayr2022} proposed a novel approach to predict the mixing pixels of an image. Defining $\textbf{x}$ as the ground truth, $\textbf{m}$ as the Boolean mask, and $\odot$ as the Hadamard product (i.e., component-wise multiplication), then, $\textbf{m} \odot \textbf{x}$ represents the known pixels and $(1-\textbf{m}) \odot \textbf{x}$ the unknown ones. Since in the forward process we can sample at any point in time using~\eqref{eq:sample_forward}, we can also sample the known regions at any time step $t$. Therefore, using~\eqref{eq:sample_forward} for the known regions and~\eqref{eq:ddpm_training} for the unknown ones, we can define:

\begin{equation}
    \textbf{x}_{t-1}^{known}\sim \mathcal{N}\left( \sqrt{\bar{\alpha}_t}\textbf{x}_0, \left( 1-\bar{\alpha}_t \right)\textbf{I} \right)
\end{equation}
\begin{equation}
    \textbf{x}_{t-1}^{unknown}\sim \mathcal{N}\left( \mu_\theta\left( \textbf{x}_t, t \right),  \Sigma_\theta\left( \textbf{x}_t, t \right)\right)
\end{equation}
\begin{equation}
    \textbf{x}_{t-1} = \textbf{m} \odot \textbf{x}_{t-1}^{known} + (1-\textbf{m}) \odot \textbf{x}_{t-1}^{unknown}.
\end{equation}

Thus, we sample $\textbf{x}_{t-1}^{known}$ using the known pixels of the given image and only sample $\textbf{x}_{t-1}^{unknown}$ from the pre-trained diffusion model. Then, we combine them to generate $\textbf{x}_{t-1}$ using the mask operation.

\section{Results}
To ensure that all events had the same length, for simplicity, a fixed grid between 7 and 0 days to the \gls{TCA} was built. Then, interpolation was performed between the data points. Finally, all series were padded on the left and right to fill in the remaining missing values.

Figure~\ref{fig:forward_diffusion_process} shows an example of applying the forward diffusion process to one of the events in the training set.

\subsection{Uncertainty Evolution Generation: Sampling from the Trained Diffusion Model}
In this study, we employed a U-Net~\cite{Ronneberger2015} architecture with 1D convolutional neural networks~\cite{Kiranyaz2021} to train the unconditional diffusion model mentioned in~\eqref{eq:ddpm_training}.

\begin{figure}[htb!]
    \centering
    \includegraphics[width=0.45\textwidth]{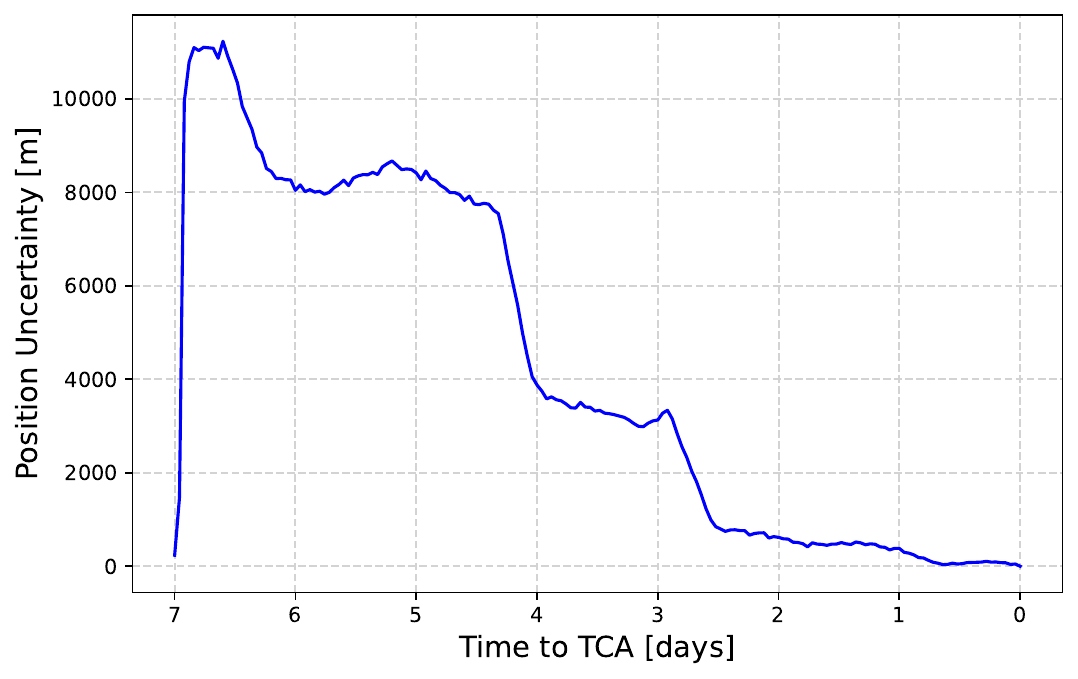}
    \caption{Generated sample from the trained diffusion model. As seen in Figure~\ref{fig:forward_diffusion_process}, the position uncertainty starts at zero. However, this should be ignored as it is a result of the padding performed in the training samples. Thus, in this example, the uncertainty starts at approximately 11k metres at around $6.7$ days to the \gls{TCA}.}
    \label{fig:sampling_from_trained_model}
\end{figure}

Having successfully trained such a model on the time series data, we now proceed to generate realistic samples. Figure~\ref{fig:sampling_from_trained_model} showcases a generated sample produced by the trained diffusion model, which effectively captures the underlying patterns and temporal dynamics present in the data. The generated time series exhibits characteristics similar to the one typically observed in real-life operations, demonstrating the model's ability to synthesize new data sequences that align with the statistical properties and dependencies observed during training.

\subsection{Forecasting as Inpainting} \label{sec:results}

After training the model, we used the inpainting technique to forecast multiple upcoming 1,000 conjunctions.

Table~\ref{tab:results} provides a comprehensive summary of the results obtained when predicting at two days prior to the \gls{TCA}, including the \gls{MAE} and the \gls{RMSE}, defined as
\begin{equation*}
    MAE = \frac{\sum_{i=1}^{n}\left| y_{i} - \widehat{y}_{i} \right|}{n} ,
\end{equation*}
\begin{equation*}
    RMSE = \sqrt{\frac{\sum_{i=1}^{n}\left( y_{i} - \widehat{y}_{i} \right)^2}{n}} ,
\end{equation*}
where $n$ is the number of predicted samples, $y_{i}$ is the real position uncertainty value and $\widehat{y}_{i}$ is the predicted one. Both metrics are expressed in the same units as the variable being predicted, i.e., in metres.

Additionally, we also present a comparative analysis with a state-of-the-art model, the Transformer~\cite{Vaswani2017}, and a baseline solution. The baseline relies on a real-life operations assumption in which the most recent value prior to forecasting represents the most informed knowledge available, and consequently, it extrapolates the remaining predicted values by maintaining the last observed step's value.

\begin{table}[!htb]
\renewcommand{\arraystretch}{1.3}
\caption{Obtained results when predicting at two days prior to the TCA.}
\label{tab:results}
\centering
\begin{tabular}{ccc}
\hline
\textbf{Model}        &  \textbf{MAE [m]}    &  \textbf{RMSE [m]}   \\ \hline
{Baseline}      & 2,772            &  5,604           \\
{Transformer}      & 1,787            &  3,525           \\
{Diffusion Model}    & \textbf{930}   &  \textbf{2,989}             \\  \hline
\end{tabular}
\end{table}

It is important to note that for both the Transformer and diffusion models, a standardized forecasting approach was adopted due to the inherent unpredictability of when actual \glspl{CDM} might exist. Thus, forecasting was conducted over a fixed grid with a step size equivalent to one hour. The subsequent evaluation was then limited to true samples residing in the proximity of specific grid steps, ensuring a fair and meaningful assessment of the models' predictive capabilities.

The analysis of the results highlights distinct performance trends among the evaluated models. 

Notably, the Baseline solution exhibited inferior performance in comparison to both the diffusion and Transformer models. This discrepancy underscores the limitations of relying solely on the continuity of the last observed value for forecasting, highlighting the need for more advanced predictive techniques to improve the safety and effectiveness of spacecraft operations.

In the case of the diffusion and Transformer models, it can be seen that our proposed solution outperformed the Transformer performance. The diffusion model demonstrated its power in generating forecasts that aligned well with the true data points, exhibiting superior levels of accuracy. This improvement in performance reaffirms the effectiveness of our proposed model in capturing temporal dependencies and predicting future trends when compared with the well-established Transformer architecture.

\begin{figure}[htb!]
    \centering
	\subfloat[Event A.]{
            \includegraphics[width=\linewidth]{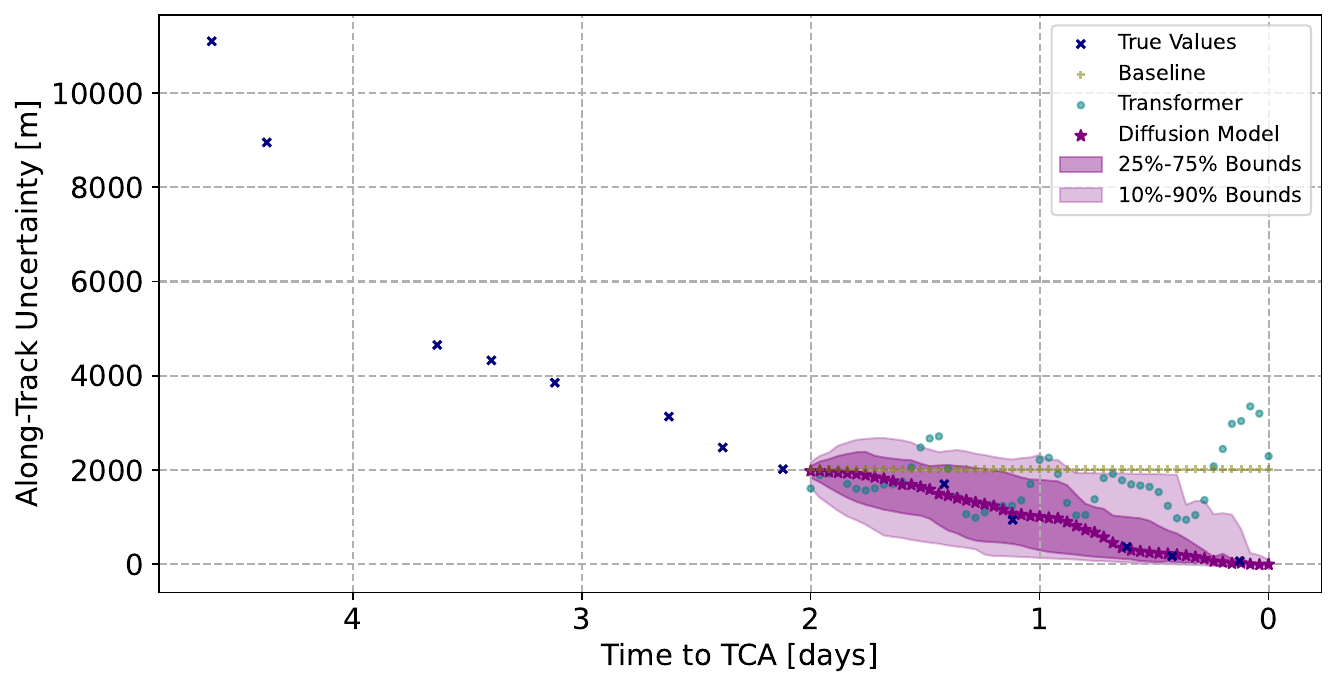}\label{fig:event_a}
        } \\
	\subfloat[Event B.]{
            \includegraphics[width=\linewidth]{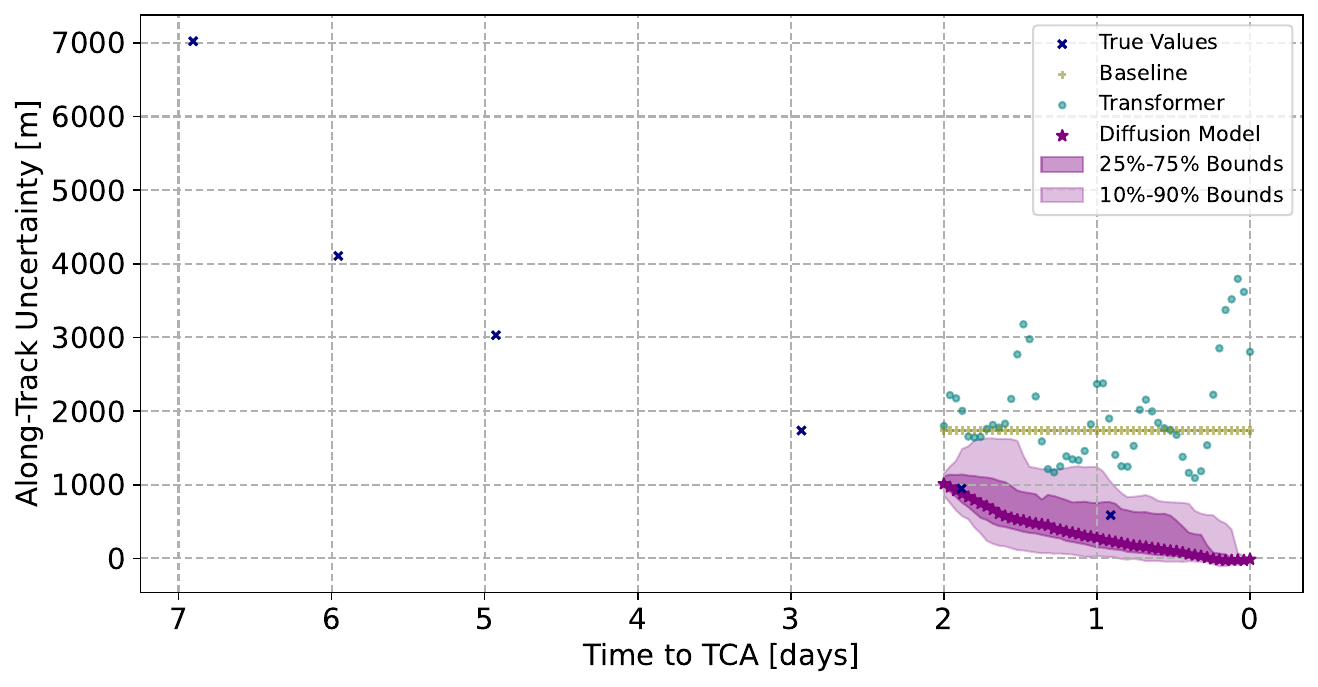}\label{fig:event_b}
        }
    \caption{Predicted time series trends when forecasting the position uncertainty at two days prior to the TCA. The dark blue crosses represent the ground truth (observed data), the green plus shows the baseline solution, the green dots describe the Transformer predictions, the purple stars depict the predictions made by the proposed diffusion model, and the shaded regions highlight the model's uncertainty quantification. In both Figures, the diffusion model exhibits a remarkable ability to capture the intricate trend of the data.}
    \label{fig:events}
\end{figure}

To substantiate this further, Figure~\ref{fig:events} showcases a selection of predicted values from both models. This selection includes not only the samples in proximity to the true values but also emphasises the trend of the predicted values.

As can be seen in the examples presented, the proposed diffusion model captures the intricate trend of the data, encompassing its fluctuations and variations.

In the first example, Figure~\ref{fig:event_a}, we examine the model's prowess when forecasting is conducted with a robust dataset, encompassing a considerable number of historical \glspl{CDM}. This situation emulates a well-informed forecasting environment, where an abundance of historical data contributes to predictive accuracy. As depicted in Figure~\ref{fig:event_a}, the diffusion model excels in generating predictions that closely align with the true values. Impressively, the model's predicted values not only stay within the uncertainty regions but also encapsulate the nuanced fluctuations and patterns, effectively capturing the underlying trend of the data.

On the other hand, the second example (Figure~\ref{fig:event_b}) introduces a challenging context, focusing on forecasting with a minimal input sequence containing only four \glspl{CDM}. This scenario emulates scenarios where data of a given conjunction is scarce, and accurate forecasts must be derived from limited information. Figure~\ref{fig:event_b} showcases the diffusion model's performance in this context, with predicted values and corresponding uncertainty regions once again enveloping the true values. The model's ability to accurately capture the prevailing trend remains evident, even within the constraints of a sparse conjunction.

Importantly, a comparative evaluation also reveals distinct tendencies in the predictions of the Transformer model. Particularly noticeable is the sinusoidal shape characterizing the predicted trend, as opposed to the more realistic patterns captured by the diffusion model. 
This observation highlights an essential distinction, as the diffusion model's predictions exhibit a remarkable consistency with the data's actual behaviour, further underlining its capacity to uncover and replicate authentic temporal patterns.

However, it is important to acknowledge the limitations that occasionally happen. 

\begin{figure}[htb!]
    \centering
        \subfloat[Event C.]{
            \includegraphics[width=\linewidth]{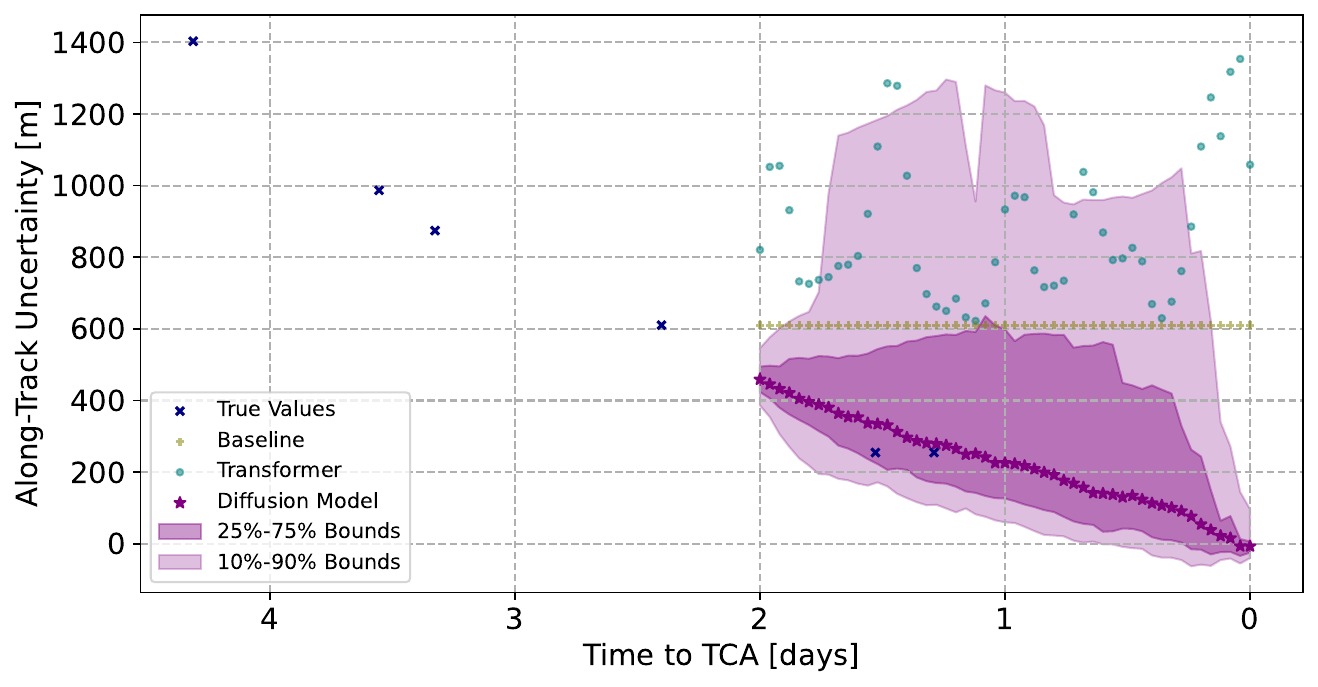}\label{fig:event_c}
        } \\
	\subfloat[Event D.]{
            \includegraphics[width=\linewidth]{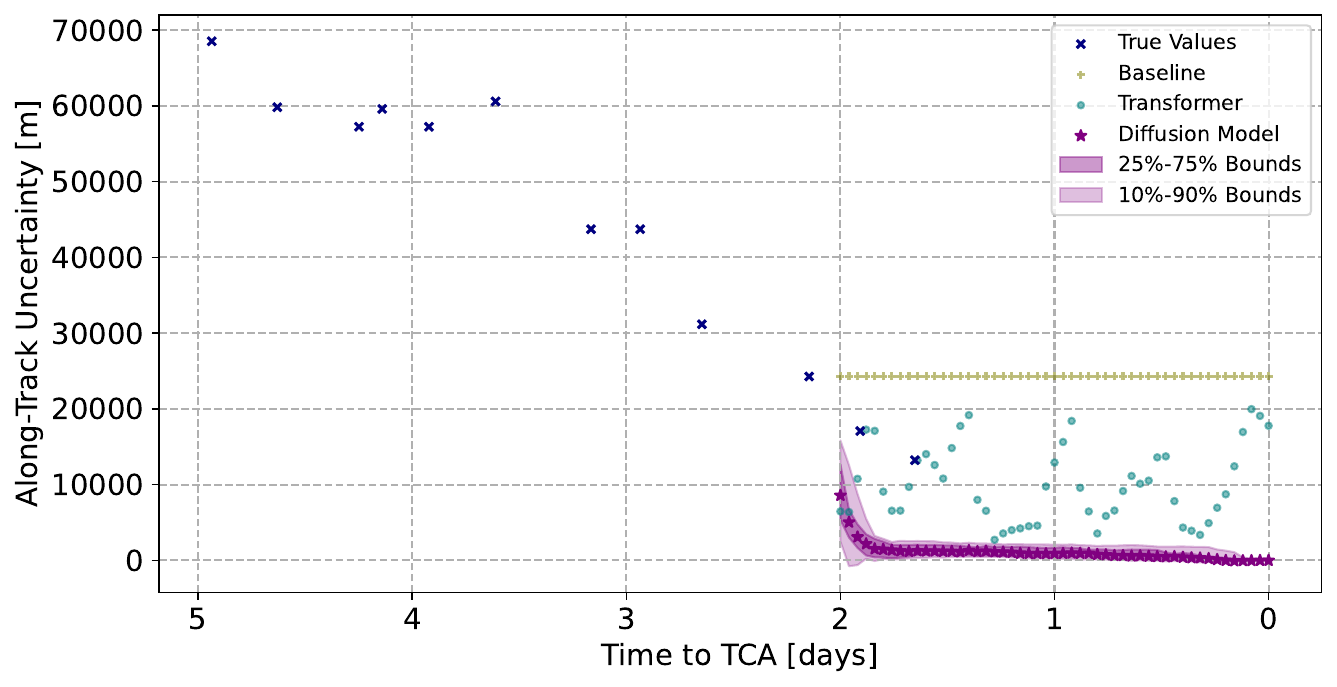}\label{fig:event_d}
        }
    \caption{Predicted time series trends when forecasting the position uncertainty at 2 days prior to the TCA. Figure~\ref{fig:event_c} showcases the scenario where the predicted values have very good performance, but the associated uncertainty noticeably deviates from the underlying trend. Figure~\ref{fig:event_d} serves as an example of the model struggling to capture the underlying trend.}
    \label{fig:events2}
\end{figure}

Figure~\ref{fig:event_c} showcases a scenario where the predicted values have very good performance, but the associated uncertainty noticeably expands, deviating substantially from the underlying trend. This scenario highlights that although model predictions may closely align with the true values, the reliability conveyed by uncertainty estimation is still dynamic and context-dependent.

In contrast, Figure~\ref{fig:event_d} presents one of the worst performing events and serves as an example of the model struggling to capture the underlying trend, evident through the pronounced steep slope exhibited in the initial predicted values. This pattern emerges particularly when dealing with conjunctions with higher covariance values before the prediction cutoff. This example underscores the model's sensitivity to certain data dynamics.

\section{Computational Trade-offs and Predictive Power}

The performance of the proposed diffusion model is not only a testament to its inherent capabilities but is also entangled with a strategic set of computational considerations.

As discussed in Section~\ref{sec:results}, the model's ability to accurately forecast time series trends is evident, as showcased by the forecasting examples presented (Figure~\ref{fig:events}). However, it is crucial to delve into the underlying factors that contribute to the model's performance and explore how variations in key parameters can further enhance its predictive power.

The forecasting examples provided were obtained using a model trained with 50 diffusion steps for a limited number of epochs. This choice was influenced by computational constraints, as training a model with an extensive number of diffusion steps and epochs would be resource-intensive. Despite these limitations, the model's forecasts exhibited a good level of accuracy, capturing the underlying trends of the data.

Furthermore, the predicted values were generated with a sample size of 32 for each example. This relatively modest sample size accounts for the uncertainty regions depicted in the forecasting plots presented in Figure~\ref{fig:events} and Figure~\ref{fig:events2}. It is important to acknowledge that a larger sample size could potentially yield more refined uncertainty estimates and further improve the predictive capabilities of the model, particularly in scenarios similar to the one shown in Figure~\ref{fig:event_c}.

To unleash the full predictive potential of the proposed diffusion model, various strategies can be considered, such as increasing the number of training epochs, employing a finer granularity of diffusion steps during training or increasing the number of samples generated for predictions. Besides, exploring deeper U-Net structures may further empower the model's predictive capabilities.

It is important to emphasise that the diffusion model's performance is a dynamic interplay between these parameters and computational resources. The trade-offs between computational efficiency and forecasting accuracy require careful consideration, and the selection of strategies should align with the specific needs of the application. While the presented examples provide a robust foundation, the potential for further enhancing the model's predictive power is substantial, enforcing its potential to significantly improve the safety and effectiveness of spacecraft operations.

\section{Conclusion}
We proposed a novel machine learning approach based on diffusion models to forecast the position uncertainty of objects involved in a close encounter. By comparing the performance of our model with other state-of-the-art solutions, i.e., the Transformer, and a naïve baseline approach, we show that the proposed solution achieves a good level of accuracy, capturing the underlying trends of the data. Our solution has the potential to significantly improve the safety and effectiveness of spacecraft operations.

\section*{Acknowledgements}
This research was carried out under Project “Artificial Intelligence Fights Space Debris” Nº C626449889-0046305 co-funded by Recovery and Resilience Plan and NextGeneration EU Funds (www.recuperarportugal.gov.pt), and by NOVA LINCS (UIDB/04516/2020) with the financial support of FCT.IP.

\bibliographystyle{unsrtnat}


\end{document}